\documentclass[conference,letterpaper,twoside,final,10pt]{ieeeconf}

\usepackage{graphicx}
\usepackage{amssymb}
\usepackage{amsmath}
\usepackage{cite}
\usepackage{comment}
\usepackage{color}
\usepackage{url}
\usepackage{alltt}
\usepackage{cleveref}
\usepackage{tikz}
\usepackage{pgfplots}
\usepackage{pgf-umlsd}
\usepackage{cprotect}
\usepackage{multirow}
\usepackage{algorithm}
\usepackage[noend]{algpseudocode}
\usepackage{subfigure}
\usepackage{siunitx}
\usepackage{bigfoot}
\usepackage{paralist}
\usepackage{tabularx}
\usepackage[whole]{bxcjkjatype}
\usepackage{ifthen}
\usepackage{threeparttable}
\usepackage{nth}
\usepackage{caption}
\usepackage[keeplastbox]{flushend}
\usepackage{bm}

\usetikzlibrary{decorations.pathreplacing,angles,quotes}

\pgfplotsset{compat=1.14}
\IEEEoverridecommandlockouts

\Crefname{figure}{Fig.}{Figs.}

\def\etal{\textit{et al.}}

\newcommand{\note}[1]{\textcolor{orange}{"COMMENT" [#1]}}
\title{\LARGE \bf
Deep Visuo-Tactile Learning: \\
Estimation of Tactile Properties from Images
}

\author{
Kuniyuki Takahashi,
Jethro Tan
\thanks{All authors are associated with Preferred Networks, Inc.
{\{takahashi, jettan\}@preferred.jp}}}

\begin{document}

\setlength\floatsep{3.0mm}
\setlength\textfloatsep{4.0mm}
\setlength\intextsep{0pt}
\setlength\abovecaptionskip{0pt}

\maketitle
\thispagestyle{empty}

%%%%%%%%%%%%%%%%%%%%%%%%%%%%%%%%%%%%%%%%%%%%%%%%%%%%%%%%%%%%%%%%%%%%%%%%%%%%%%%%
%%%%%%%%%%%%%%%%%%%%%%%%%%%%%%%%%%%%%%%%%%%%%%%%%%%%%%%%%%%%%%%%%%%%%%%%%%%%%%%%
\begin{abstract}%less than 200 words
Estimation of tactile properties from vision, such as slipperiness or roughness, is important to effectively interact with the environment.
These tactile properties help us decide which actions we should choose and how to perform them. E.g., we can drive slower if we see that we have bad traction or grasp tighter if an item looks slippery.
We believe that this ability also helps robots to enhance their understanding of the environment, and thus enables them to tailor their actions to the situation at hand.
We therefore propose a model to estimate the degree of tactile properties from visual perception alone (e.g., the level of slipperiness or roughness).
Our method extends a encoder-decoder network, in which the latent variables are visual and tactile features.
In contrast to previous works, our method does not require manual labeling, but only RGB images and the corresponding tactile sensor data.
All our data is collected with a webcam and uSkin tactile sensor mounted on the end-effector of a Sawyer robot, which strokes the surfaces of 25 different materials.\footnote{Dataset is available at the following link:\\ \url{https://github.com/pfnet-research/Deep_visuo-tactile_learning_ICRA2019}}
We show that our model generalizes to materials not included in the training data by evaluating the feature space, indicating that it has learned to associate important tactile properties with images.\footnote{An accompanying video is available at the following link:\\ \url{https://youtu.be/ys0QtKVVlOQ}}
\end{abstract}
%%%%%%%%%%%%%%%%%%%%%%%%%%%%%%%%%%%%%%%%%%%%%%%%%%%%%%%%%%%%%%%%%%%%%%%%%%%%%%%%
%%%%%%%%%%%%%%%%%%%%%%%%%%%%%%%%%%%%%%%%%%%%%%%%%%%%%%%%%%%%%%%%%%%%%%%%%%%%%%%%
\section{Introduction}
\label{sec:introduction}
\begin{figure}
	\centering
    \subfigure[hard, textured]{\includegraphics[width=0.32\columnwidth]{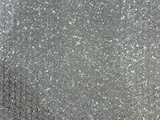}}
	\subfigure[soft, fluffy]{\includegraphics[width=0.32\columnwidth]{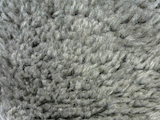}}
	\subfigure[soft/hard, coarse]{\includegraphics[width=0.32\columnwidth]{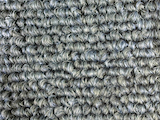}}
	\caption{Example of material surfaces and their perceived material properties through visual information.}
	\label{fig:propertyassociation}
\end{figure}

Humans are able to perceive tactile properties, such as slipperiness and roughness, through haptics~\cite{bergmann2010tactual}.
However, after adequate visual-tactile experience, they are also capable of associating such properties from only visual perception~\cite{tanaka2015investigating, yanagisawa2015effects}. 
More specifically, humans can roughly judge the \emph{degree} of a certain tactile property (e.g., the \emph{level} of slipperiness or roughness)~\cite{fleming2014visual}.
As an example, \Cref{fig:propertyassociation} shows several materials with different degrees of softness and roughness judged by ourselves, although this may be subjective to our own judgment.
Information on tactile properties can help us decide how we interact with our environment in advance, e.g., driving slower if we see that we have bad traction or grasp tighter if an item looks slippery.
Like with humans, this ability to gauge the level of tactile properties can enable robots to deal with various objects and environments in both industrial settings and our daily lives more effectively.
\begin{comment}{
We argue that judging the degree of tactile properties is of more importance than knowing what type of item we are dealing with.
In our examples, our actions when driving on a slippery road are more determined by \emph{how} slippery it is, and less by \emph{what} lies ahead on the road (though knowing what exactly lies ahead helps us determine the degree of slipperiness).
\note{This sentence is not sure or too strong. Sometimes people want to know type of material. I think we don't need this red part sentences.}
}
Like with humans, this ability to gauge the level of tactile properties can enable robots to deal with various objects and environments in both industrial settings and our daily lives more effectively.
\end{comment}

In the field of robotics and machine learning, a straightforward way to correlate vision with tactile properties is to design discrete classes per material type and to classify the images according to them.
However, an adequate number of training labels is required in order to cover a broad range of material types.
This is especially true if the image shows an unknown material type that does not appear in the training dataset.
In other words, the performance of discrete classification methods highly depends on how well the designer chooses the number and types of class labels.
Because of the wide variety of materials, which all have different tactile properties, discrete classes can not offer a sufficient resolution to judge the properties of the material well.
We argue that discrete classification is unfit for our purposes since the wide range of material types would require a large number of classes.
Moreover, we are not interested in categorizing the material type, but rather in estimating the \emph{degree} of the tactile properties.

Hence, we use an unsupervised method to represent tactile properties without using manually specified labels.
We propose a method that we call \emph{deep visuo-tactile learning} which extends a traditional encoder-decoder network with latent variables, where visual and tactile properties are embedded in a latent space.
We emphasize that this is a \emph{continuous} space, rather than a discrete one.
This method is capable of generalizing to new, unknown materials when estimating their tactile properties, based on known tactile properties.
Additionally, we only require the tactile sensor during the data collection phase and obtain a trained network model that can be used even in simulations or offline estimation, which allows for further research without purchasing or damaging tactile sensors during runtime.

The rest of this paper is organized as follows.
Related work is described in~\Cref{sec:relatedwork}, while~\Cref{sec:method} explains our proposed method.
\Cref{sec:experiments} outlines our experiment setup and evaluation settings with results presented in~\Cref{sec:results}.
Finally, future work and conclusions are described in~\Cref{sec:conclusion}.

%%%%%%%%%%%%%%%%%%%%%%%%%%%%%%%%%%%%%%%%%%%%%%%%%%%%%%%%%%%%%%%%%%%%%%%%%%%%%%%%
%%%%%%%%%%%%%%%%%%%%%%%%%%%%%%%%%%%%%%%%%%%%%%%%%%%%%%%%%%%%%%%%%%%%%%%%%%%%%%%%

\section{Related Work}
\label{sec:relatedwork}
\subsection{Development of Tactile Sensors}
Many researchers have developed tactile sensors~\cite{dahiya2013directions}, some of which have been integrated to a robotic hand to enhance manipulation.
The majority of these sensors, however, falls in either of the following two categories.
\begin{inparaenum}\item Multi-touch enabled sensors with sensing capabilities limited to one axis per cell~\cite{ohmura2006conformable,iwata2009design,mittendorfer2011humanoid,fishel2012sensing} or \item three-axis sensing enabled sensors for only a single cell~\cite{paulino2017low}\end{inparaenum}.
Two of the few exceptions are the GelSight~\cite{johnson2009retrographic, dong2017improved} and the uSkin~\cite{tomo2016modular, tomo2018uskin}.
The GelSight is an optical-based tactile sensor, which uses a camera to record and measure the deformation of its attached elastomer during contact with a surface. 
By using markers on its surface and detecting their displacements, shear force can also be measured.
While the GelSight has an impressive spatial resolution in the range of up to 30--100\,microns~\cite{ yuan2017gelsight}, the elastomer is easily damaged during contact and thus requires frequent maintenance in contact-rich manipulation tasks such as grasping~\cite{Calandra2017}.
Instead of a camera, the commercialized uSkin sensor by Tomo~\etal in~\cite{tomo2016modular, tomo2018uskin}, utilizes magnets to measure the deformation of silicon during contact by monitoring changes to the magnetic fields.
Using this method, it is able to measure both normal as well as shear forces for up to 16 contact points per sensor unit.
By additionally covering the silicon surface with lycra fabric, the durability of the sensor against friction can be enhanced to minimize maintenance.

%%%%%%%%%%%%%%%%%%%%%%%%%%%%%%%%%%%%%%%%%%%%%%%%%%%%%%%%%%%%%%%%%%%%%%%%%%%%%%%%
\subsection{Recognition through Tactile Sensing}
Research utilizing tactile sensors has grown recently as the availability and accessibility to tactile sensors has improved.
Prior to the use of deep learning-based methods in these studies, data acquired from tactile sensors were often analyzed manually in order to define hand-crafted features~\cite{yang2016tactile}, or were only used as a trigger for a certain action~\cite{yamaguchi2016combining}.
However, such methods may not scale well as technology for tactile sensing advances to provide e.g., higher resolution and larger amount of data, or whenever the task complexity grows.
By utilizing learning methods, especially deep learning, tasks involving high-dimensional data such as image recognition~\cite{he2016deep} and natural language processing~\cite{conneau2016very} which were too difficult to process before can now be processed.
Soon afterwards, deep learning methods also found their way to applications where tactile sensing is involved~\cite{schmitz2014tactile, baishya2016robust, yuan2017connecting, gao2016deep}.
Many of these studies, however, deal with the classification problem in order to e.g., recognize objects inside a robotic hand~\cite{schmitz2014tactile}, recognize materials~\cite{baishya2016robust, yuan2017connecting} and properties~\cite{gao2016deep} from touch and image.
Yuan~\etal\cite{yuan2017shape} estimated object hardness as a continuous value using tactile sensor through supervised learning.
However, we argue that their method would be difficult to scale to different tactile properties due to the need of designing each tactile property manually.

A different use case is shown in~\cite{Calandra2017} where Calandra~\etal  utilizes deep reinforcement learning and combined data acquired from a tactile sensor and images as network input to grasp objects, which improved their success rate in grasping experiments.
Similar to previous studies, however, they also require the tactile sensor to be present during task execution.
Our work differs from previous works in that we only make use of the tactile sensor while collecting data to finally train our neural network.
Afterwards, no tactile sensor is needed to estimate the tactile properties from input images.

We also note that there are other related studies on recognition of materials without utilizing tactile sensors, such as~\cite{bell2015material, schwartz2017visual}.
However, they primarily focus on either categorization or classification of material types like e.g., stone, wood, fabric, etc., which differs from our goal to estimate tactile properties as well as their degree in this work.
%%%%%%%%%%%%%%%%%%%%%%%%%%%%%%%%%%%%%%%%%%%%%%%%%%%%%%%%%%%%%%%%%%%%%%%%%%%%%%%%
%%%%%%%%%%%%%%%%%%%%%%%%%%%%%%%%%%%%%%%%%%%%%%%%%%%%%%%%%%%%%%%%%%%%%%%%%%%%%%%%
\begin{figure*}[tb]
  \centering
  \includegraphics[width=1.6\columnwidth]{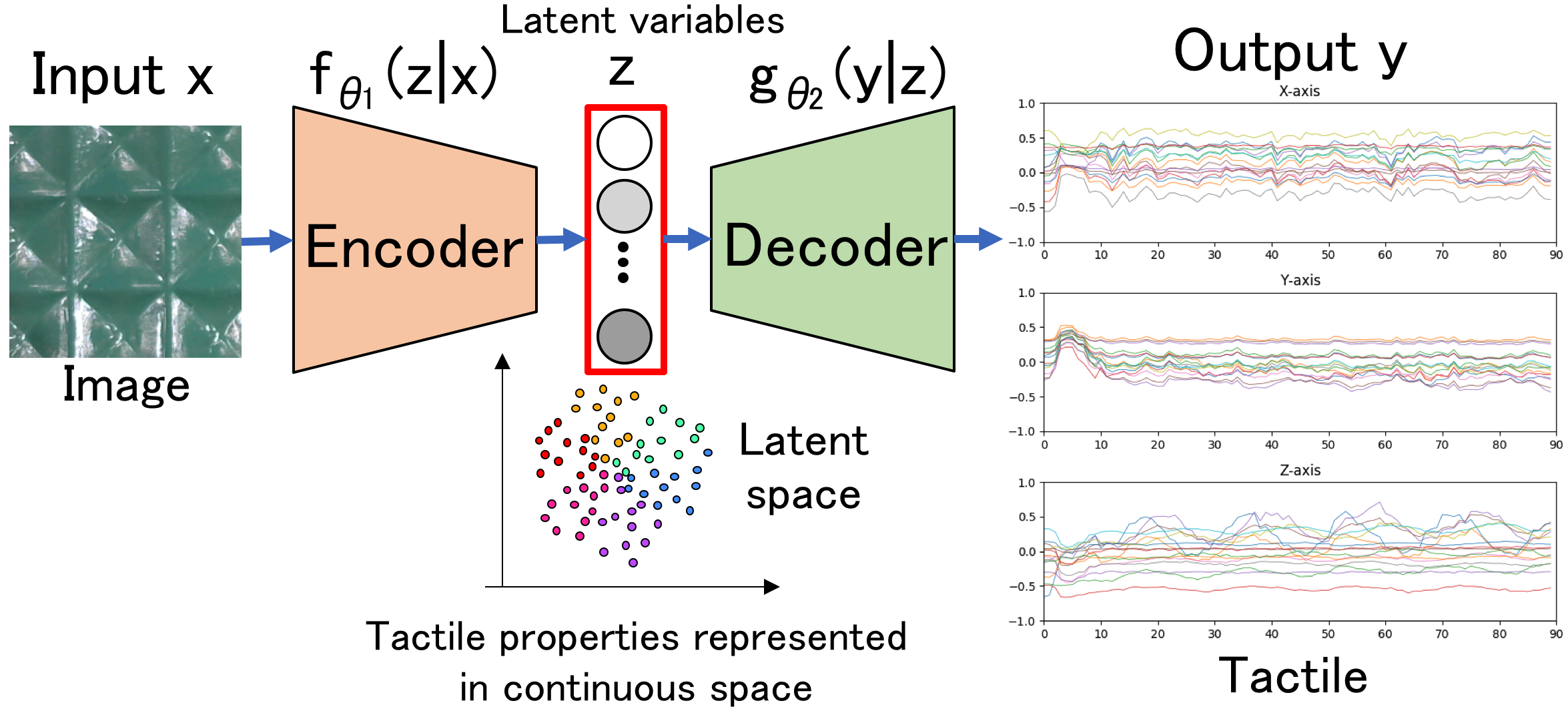}
  \caption{
    Proposed network architecture for deep visuo-tactile learning composed of encoder-decoder layers and latent variables.
    Input is texture image of material and, output is the tactile data contains measured forces by a tactile sensor in the x, y, and z axes.
    After training, latent variables would contain tactile properties of materials correlating images with tactile sense.}
	\label{fig:architecture}
\end{figure*}
%%%%%%%%%%%%%%%%%%%%%%%%%%%%%%%%%%%%%%%%%%%%%%%%%%%%%%%%%%%%%%%%%%%%%%%%%%%%%%%%
%%%%%%%%%%%%%%%%%%%%%%%%%%%%%%%%%%%%%%%%%%%%%%%%%%%%%%%%%%%%%%%%%%%%%%%%%%%%%%%%
\section{Deep Visuo-tactile learning}
\label{sec:method}
We propose a method for deep visuo-tactile learning to estimate tactile properties from images by associating tactile information with images.
\Cref{fig:architecture} shows our design of such a network.
We aimed to design a network with a structure that is as simple as possible, but still sufficient for our purposes.
We expect that increased complexity of the network architecture by e.g., using variational auto-encoder (VAE) and recurrent neural networks will mainly influence the accuracy and how tactile properties are represented as features, but that the results remain analogous. Complex models usually have the ability to learn more complex representations and larger datasets, but our contribution can be shown using simpler models, hence our decision.

Our proposed network consists of 2D convolution layers for encoding, 3D deconvolution layers for decoding, and a multi layer perceptron (MLP) as hidden layers between the encoder $f_{\theta}$ and decoder  $g_{\theta}$.
Convolutional neural networks (CNNs) are neural networks that convolve information by sliding a small area called a filter.
2D convolutions are often used in CNNs for static images with the purpose of sliding the filter along the image plane.
For images with time series information (e.g., a video), 3D convolutions are used instead to convolve information by sliding a small cubical region along 3D space~\cite{ji20133d}.
Our network outputs a time series sequence of tactile data consisting of applied forces and shear forces, while the input is an edge extracted image from the RGB image to prevent correlation to colors.
The latent variables $z$ are calculated with training data $D = \{(x_{1},y_{1})...,(x_{n},y_{n})\}$ to minimize the cost function $L$ as follows:

\begin{equation}
z=f_{{\theta}_{1}}(x) = s_{enc}(\bm{W}_{1}x+b_{1}),
\label{eq:encode_eq}
\end{equation}

\begin{equation}
y^{\prime}=g_{{\theta}_{2}}(z) = s_{dec}(\bm{W}_{2}z+b_{2}),
\label{eq:decode_eq}
\end{equation}

\begin{equation}
\min_{{\theta}_{1},{\theta}_{2}}\frac{1}{n}\sum_{i=1}^{n} L(y_{i},g_{{\theta}_{2}}(f_{{\theta}_{1}}(x_{i}))),
\label{eq:cost_eq}
\end{equation}

where  $z$ are the latent variables, $s_{enc}$ and $s_{dec}$ are the activation functions for the encoder and decoder, respectively, and $\theta = (\bm{W}, b)$ are the parameters to be trained.
$y {\in} {\bm{R}}^{d}$ is the expected output, and $y^{\prime} {\in} {\bm{R}}^{d}$ is the inferred output from input $x {\in} {\bm{R}}^{d}$.

After training, $z$ will hold visuo-tactile features that can be used to correlate the input images to the time series tactile data.
We then map the embedded input to the latent space spanned by these variables; the coordinates of the embeddings in this space will represent the material's degree of the tactile property represented by the latent variable.
However, we remind the reader that we do not focus on inferring the tactile time series data as output from the input images.
Rather, we attempt to estimate the level of tactile properties, which can now be done by extracting the latent variables from the trained network.
The reason for not directly using the values from the inferred time series data is that they are too sensitive to contact differences in e.g., the posture used to initiate the contact, the movement speed during contact, and the wear condition of the contact surface.

%%%%%%%%%%%%%%%%%%%%%%%%%%%%%%%%%%%%%%%%%%%%%%%%%%%%%%%%%%%%%%%%%%%%%%%%%%%%%%%%
%%%%%%%%%%%%%%%%%%%%%%%%%%%%%%%%%%%%%%%%%%%%%%%%%%%%%%%%%%%%%%%%%%%%%%%%%%%%%%%%

\section{Experiment Setup}
\label{sec:experiments}

\subsection{Hardware Setup}
\label{ssec:experimentsetup}
\begin{figure}
  \centering
  \includegraphics[width=1.0\columnwidth]{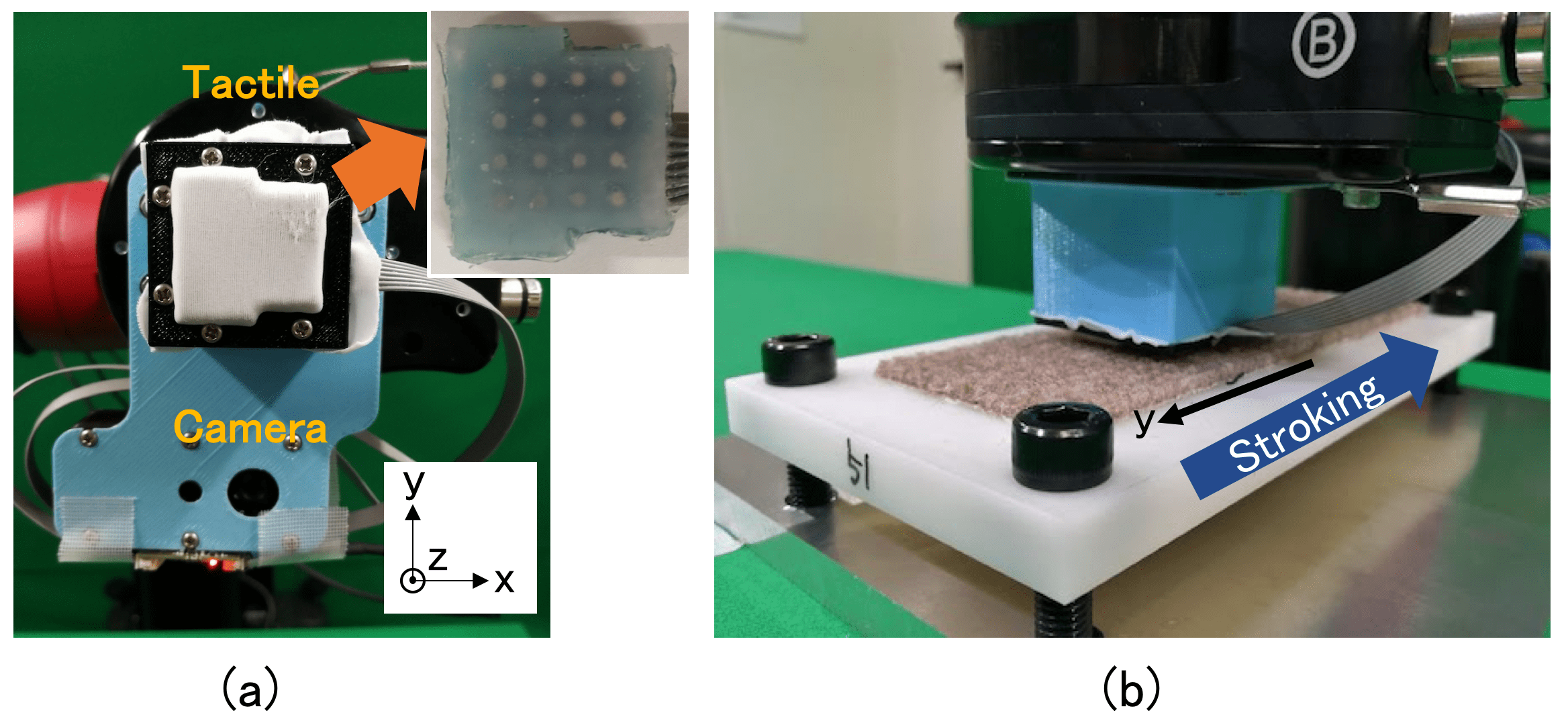}
  \caption{Setup used in our experiments: (a) custom printed end-effector with both a tactile skin sensor and a HD web camera, and (b) the Sawyer robot stroking a material sample to the minus y-axis direction.}
  \label{fig:setup}
\end{figure}
\subsubsection{Tactile sensor}
The uSkin tactile sensor we use~\cite{tomo2016modular,tomo2018uskin} consists of 16 taxels in a $4\times4$ square formation and is capable of measuring applied pressing forces and shear force in the x, y, and z axes as well as temperature (\Cref{fig:setup}(a) shows the coordinate system of the tactile sensor).
For our experiments, we only use the raw values of the pressure readings $x,y,z \in [0, 65535]$ on each of the taxels, which are configured to sample at 100\,Hz.
According to the manufacturers, the uSkin can handle pressing forces up to 40.0\,N in its z-axis.
However, both shear forces (i.e., in x and y axes) are limited to about 2.0\,N due to the physical limits of the silicone layer.
Applying an excess amount of force results in tearing the silicon layer from the sensor's PCB forcing maintenance of the entire sensor.
To prevent this from happening, we have covered all surfaces of the sensor with lycra fabric as suggested by the manufacturers.
%%%%%%%%%%%%%%%%%%%%%%%%%%%%%%%%%%%%%%%%%%%%%%%%%%%%%%%%%%%%%%%%%%%%%%%%%%%%%%%%

\subsubsection{Materials}
\begin{figure}
    \includegraphics[width=1.0\columnwidth]{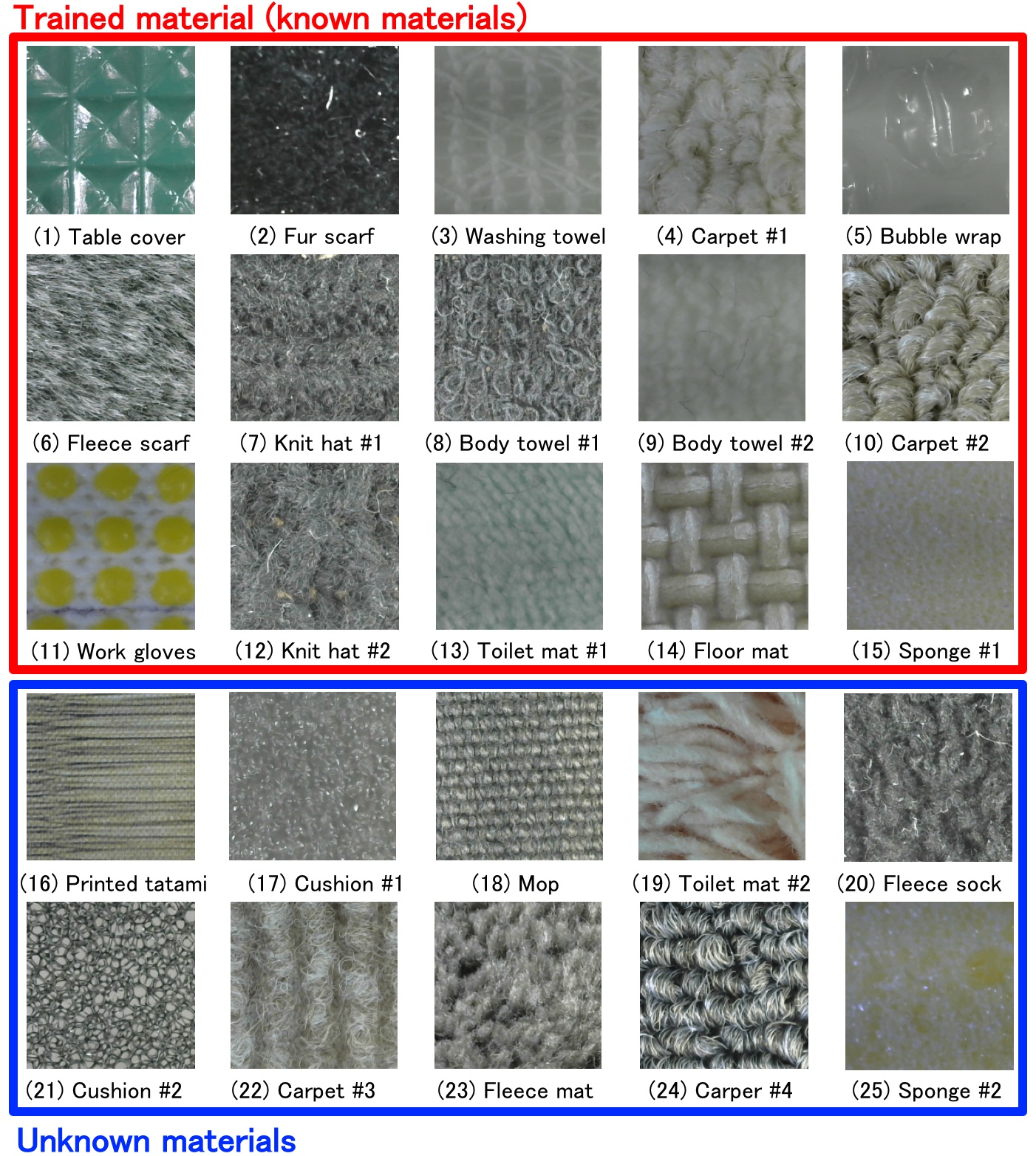}
    \caption{Trained materials (red) and unknown materials (blue) with their corresponding names included our dataset.}
    \label{fig:dataset}
\end{figure}

For the materials, we have prepared 50x150\,mm samples of 25 materials with different textures and rigidity that can be obtained off the shelf from a hardware store, see~\Cref{fig:dataset}.
15 of these materials are used for training, while the remaining 10 were used to evaluate our trained network as unknown materials.
To normalize the experiments between each material and simplify the process of our data collection, we have glued each of the samples to their own PVC plate (See~\Cref{fig:setup}(b)).
The PVC plates themselves are held on to their position per experiment by bolts that are inserted to a heavy metallic plate on top of the experiment table.

%%%%%%%%%%%%%%%%%%%%%%%%%%%%%%%%%%%%%%%%%%%%%%%%%%%%%%%%%%%%%%%%%%%%%%%%%%%%%%%%
\subsubsection{Sawyer}
To conduct our experiments, we make use of a Sawyer 7-DOF robotic arm with a custom 3D-printed end-effector on which the uSkin tactile sensor and a Logitech C310 HD camera are mounted (See~\Cref{fig:setup} (a)).
The uSkin sensor is connected to a PC running Ubuntu 16.04 with ROS Kinetic, which also controls all other hardware components including the robot controller.
%%%%%%%%%%%%%%%%%%%%%%%%%%%%%%%%%%%%%%%%%%%%%%%%%%%%%%%%%%%%%%%%%%%%%%%%%%%%%%%%
\subsection{Data Collection}
For data collection, the following process is repeated ten times per material by the robot.
\begin{enumerate}
	\item Move to a fixed initial position
	\item Detect material surface: move down from a fixed initial height until force threshold $\theta_{Fz}$ \num{5.0}\.N has been reached
	\item Capture image: move up \num{1.6e-2}\,m from detected material surface and take a picture
	\item Move back to material surface and start capturing data from tactile sensor
	\item Stroke material: move \num{3.0e-2}\,m with constant velocity \num{2.0e-3}\,m/sec in positive y-axis direction while tactile sensor makes contact with material surface
\end{enumerate}

\algdef{SE}[DOWHILE]{Do}{doWhile}{\algorithmicdo}[1]{\algorithmicwhile\ #1}%
\begin{algorithm}
\caption{Material Stroking Process}
\label{alg:strokingprocess}
\begin{algorithmic}
\State $q_{angles} \gets \textsc{init}$ \Comment{Send robot to init position.}
\For{$\rm{\textsc{iter}}=1$ to $10$}
	\State $q_{angles} \gets \textsc{home}$ \Comment{Send robot to home position.}
	\While{$F_{z} > \theta_{Fz}$}
		\State $q_{pos}(z) \gets q_{pos}(z) - \Delta z_{down}$ \Comment{Move tool down.}
    \EndWhile
	\State $q_{pos}(z) \gets q_{pos}(z) + \Delta z_{cam}$ \Comment{Move tool up.}
    \State \textsc{CaptureImage()}
    \While{$F_{z} > \theta_{Fz}$}
		\State $q_{pos}(z) \gets q_{pos}(z) - \Delta z_{down}$ \Comment{Move tool down.}
    \EndWhile
    \While{not $(F_{x} > \theta_{Fx}$ and $d_{stroked} < d_{stroke})$}
		\State $q_{pos}(x) \gets q_{pos}(x) - \Delta x_{stroke}$ \Comment{Stroke material.}
    \EndWhile
\EndFor
\end{algorithmic}
\end{algorithm}

After data collection, we process all data to obtain our training data by doing the following.
We first calibrate each acquired tactile sequence using its first 50 time steps.
Afterwards, we normalize all remaining values to be between -1 and 1 and sample down each sequence of 900 time steps to 90 steps.
Moreover, we perform rotations and croppings (from $640\times480$\,pixels to pieces of $200\times200$\,pixels) covering various areas to the obtained images.
By doing this, we augment our data by 64 times per material and obtain a total of 960 samples of image-tactile pairs.
Furthermore, we extract the edges from the RGB images of the materials with normalized pixel values between -1 and 1, because we reason that touch sense does not depend on material colors, and performing this preprocessing enables us to train our network with less data.
For training, we use eight out of the ten collected images and tactile sequences.
The remaining two image-tactile sequence pairs were split for validation and testing, respectively.
\begin{comment}
%    For research reproducibility, our dataset containing the full list of names of all the materials as well as training data will be made available online.
\end{comment}
%%%%%%%%%%%%%%%%%%%%%%%%%%%%%%%%%%%%%%%%%%%%%%%%%%%%%%%%%%%%%%%%%%%%%%%%%%%%%%%%
\subsection{Network Hyper-parameters \& Training}
\label{ssec:networkarchitecture}
The architecture of our network model with four 2D convolutional, four 3D convolutional, and two full-connected MLPs to perform deep visuo-tactile learning is shown in~\Cref{fig:architecture} as described in~\Cref{sec:method}.
More details on the network parameters are shown in~\Cref{tab:networkdesign}.
For all layers except last layer in the network, we make use of batch normalization.
For training, we use mean squared error as cost function, and Adam~\cite{kingma2014adam} as optimizer with $\alpha=$\num{1e-3} and batch size of 15.
All our network experiments were conducted on a machine equipped with 128\,GB RAM, an Intel Xeon E5-2623v3 CPU, and a GeForce GTX Titan X with 12GB resulting in about 1.5 hours of training time.

\begin{table}
\vspace{5mm}
\centering
\begin{threeparttable}
\caption{Network Design\tnote{1}}
\label{tab:networkdesign}
\begin{tabular}{|c|c|c|c|c|c|c|c|}
\hline
 & Layer & In & Out & \shortstack{Filter\\size} & Stride & Padding & \shortstack{Activation\\function}\\
\hline\hline
 \parbox[t]{2mm}{\multirow{4}{*}{\rotatebox[origin=c]{90}{Encoder}}}
 		& \nth{1} & 1  & 32 & (8,8) & (2,2) & (0,0) & ReLu\\
        & \nth{2} & 32 & 32 & (8,8) & (2,2) & (0,0) & ReLu \\
        & \nth{3} & 32 & 32 & (4,4) & (2,2) & (0,0) & ReLu \\
        & \nth{4} & 32 & 32 & (4,4) & (2,2) & (0,0) & Tanh \\
\hline
 \parbox[t]{2mm}{\multirow{4}{*}{\rotatebox[origin=c]{90}{Decoder}}}
 		& \nth{1} & 1  & 32 & (1,1,3) & (1,1,1) & (0,0,0) & ReLu \\
        & \nth{2} & 32 & 32 & (1,1,3) & (1,1,2) & (0,0,0) & ReLu \\
        & \nth{3} & 32 & 32 & (2,2,4) & (1,1,2) & (0,0,3) & ReLu \\
        & \nth{4} & 32 & 3  & (2,2,4) & (1,1,2) & (1,1,2) & Tanh \\
\hline
\end{tabular}
\begin{tablenotes}
	\item[1] \scriptsize {For the hidden layer between encoder and decoder, we use two MPLs with 4 and 160 neurons with ReLu as activation function, respectively.}
\end{tablenotes}
\end{threeparttable}
\end{table}
%%%%%%%%%%%%%%%%%%%%%%%%%%%%%%%%%%%%%%%%%%%%%%%%%%%%%%%%%%%%%%%%%%%%%%%%%%%%%%%%
%%%%%%%%%%%%%%%%%%%%%%%%%%%%%%%%%%%%%%%%%%%%%%%%%%%%%%%%%%%%%%%%%%%%%%%%%%%%%%%%

\section{Results}
\label{sec:results}
\subsection{Tactile Sequences Data}
\label{ssec:results1}
\begin{figure}
  \centering
  \includegraphics[width=1.0\columnwidth]{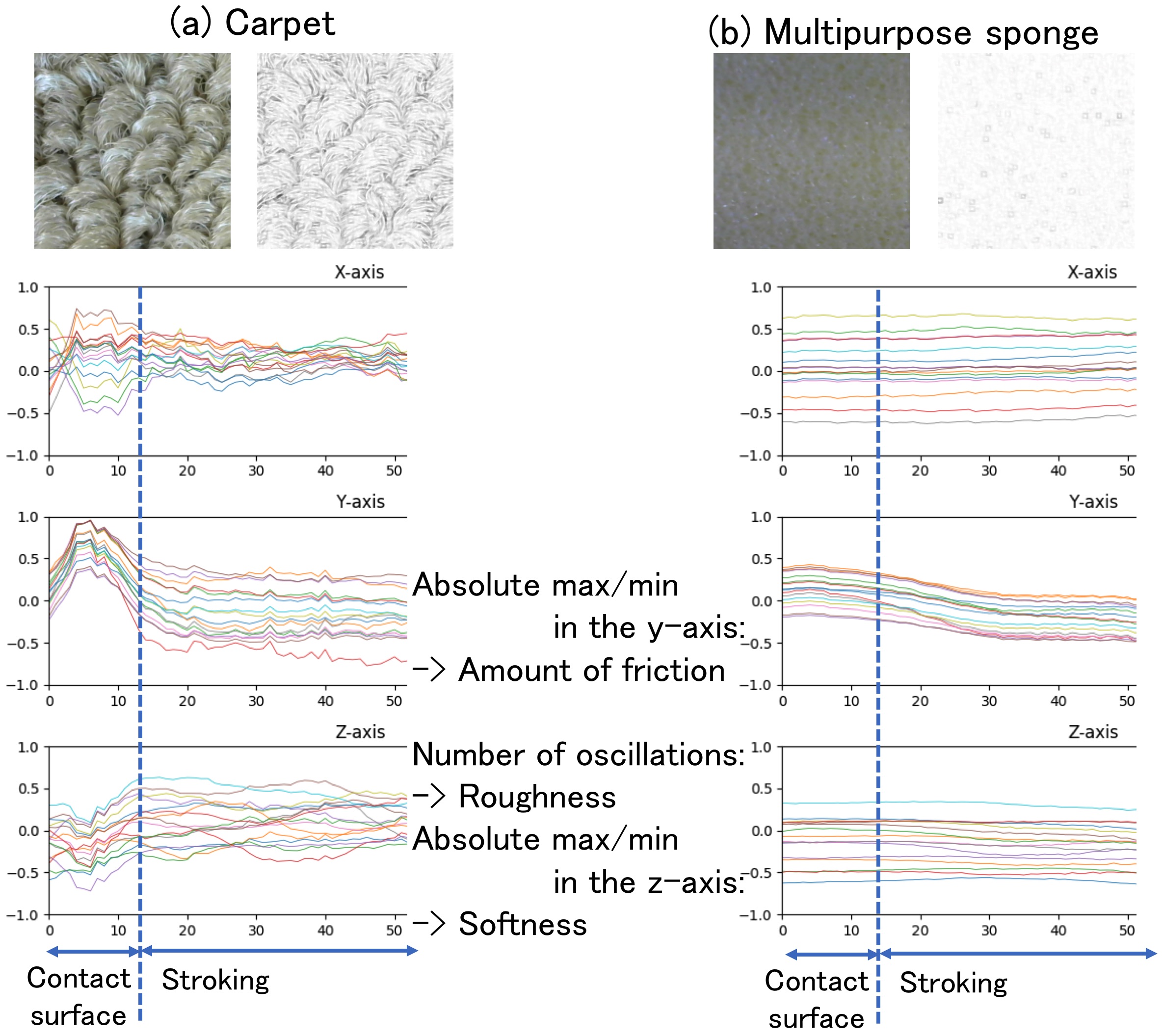}
  \caption{Examples of cropped colored and preprocessed images with their corresponding tactile sequences from the material samples: (a) carpet, and (b) multipurpose sponge.
  }
  \label{fig: example of image and tactile}
\end{figure}

We first show example plots of tactile sequence data with forces in the x, y and z axes for all the 16 sensor taxels (\Cref{fig: example of image and tactile}).
The material shown in~\Cref{fig: example of image and tactile}~(a) is a patch of carpet, while~\Cref{fig: example of image and tactile}~(b) shows a piece of a multipurpose sponge.
We expect that the y-axis values contain information on friction between the end-effector and the material due to the applied shear forces while stroking.
We also observe a waveform in the z-axis graph of the carpet due to its uneven surface.
For the sponge on the other hand, we see that relative changes in forces are small in comparison to the carpet during the stroking movement, because most of the forces are damped by the softness of the sponge.
Therefore, we believe that the z-axis embeds not only information on roughness, but also softness of a material surface.
In a similar fashion, other properties of various materials expressed as numerical values might be embedded inside the acquired tactile information.

%%%%%%%%%%%%%%%%%%%%%%%%%%%%%%%%%%%%%%%%%%%%%%%%%%%%%%%%%%%%%%%%%%%%%%%%%%%%%%%%
\subsection{Estimation of Tactile Properties}
\begin{figure}
	\centering
	\subfigure[]{\includegraphics[width=0.9\columnwidth]{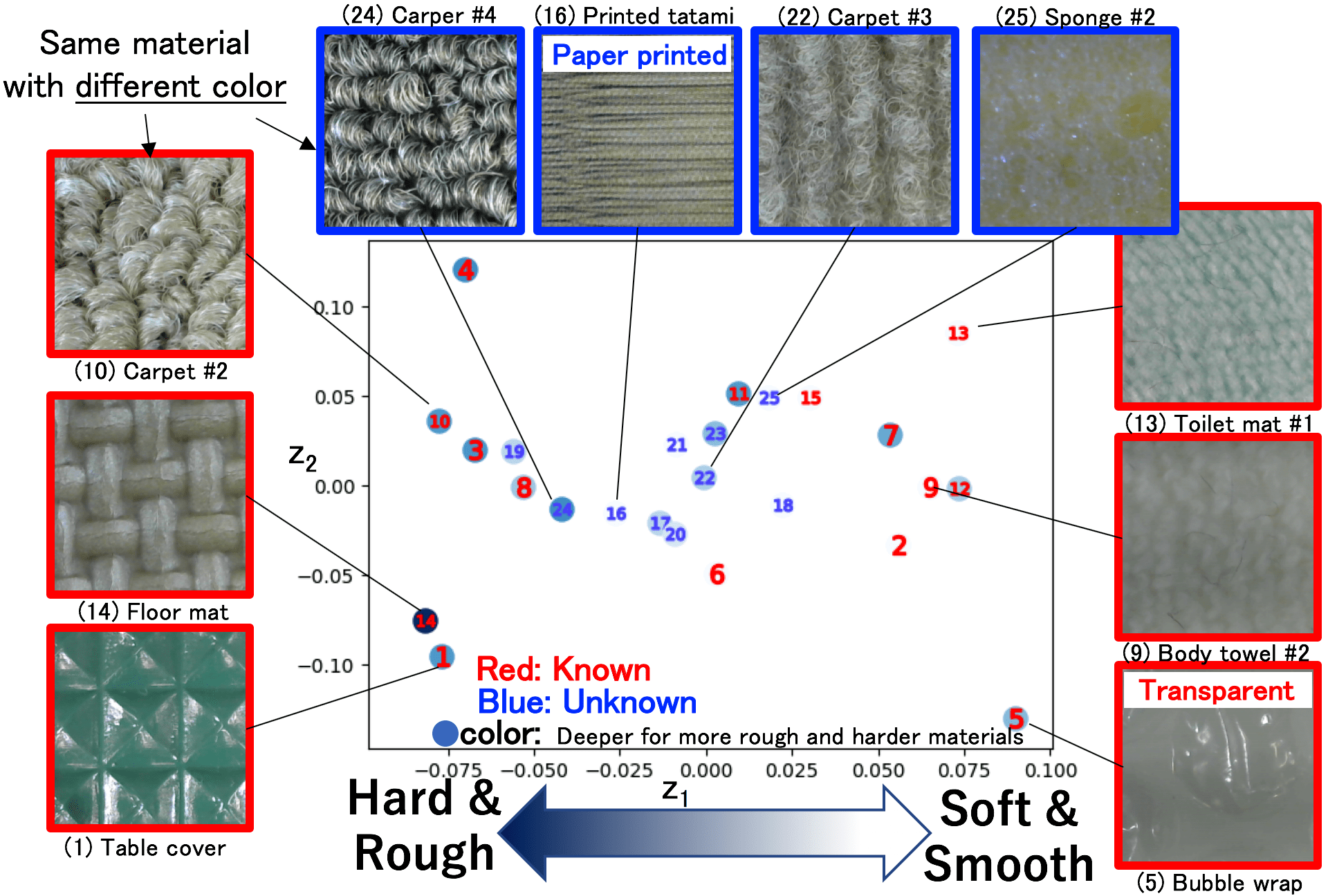}}
	\subfigure[]{\includegraphics[width=1.0\columnwidth]{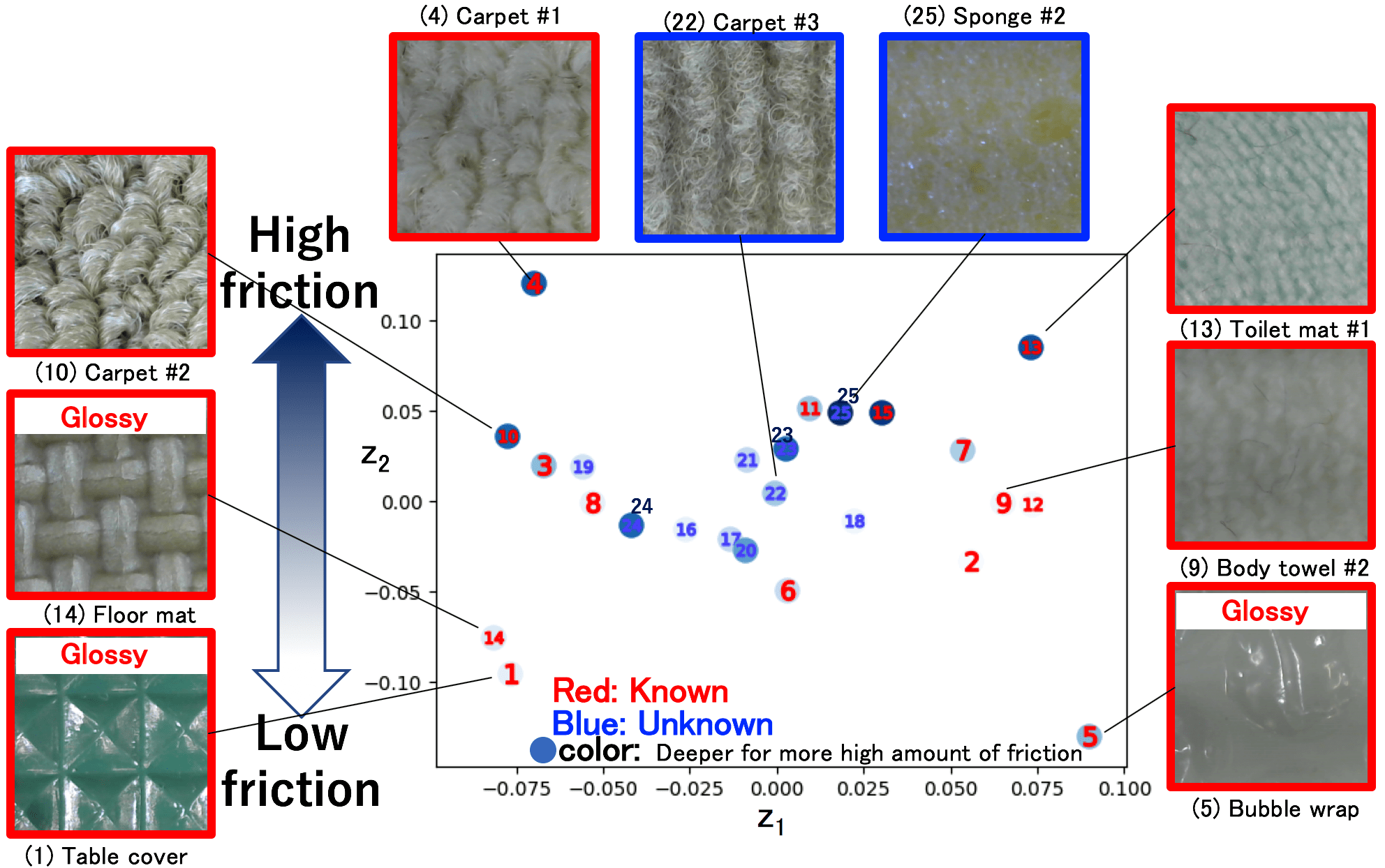}}
	\caption{Visualization of tactile properties of (a) softness and roughness, and (b) friction from latent spaces of the hidden layer.}
	\label{fig:visualization}
\end{figure}

Here, we present the results of estimated tactile properties in the latent space.
After training with the 15 known materials shown in~\Cref{fig:dataset}, we let our network infer tactile properties with both known and 10 additional unknown materials.
The tactile properties for all these materials are represented in four latent variables of the hidden layer.
\Cref{fig:visualization} shows the latent space of two of those latent variables in the hidden layer.
We have, to the best of our ability, analyzed the remaining two latent variables, but infer that the information they seem to represent are too diverse to analyze tactile properties.
Known materials used during training are represented with their corresponding red-colored numbers as found in~\Cref{fig:dataset},
while unknown materials are represented in their corresponding blue-colored numbers.

To qualitatively evaluate the results of how tactile properties are represented in the latent space, we calculate the values for roughness, hardness, and friction for each material as described in~\Cref{ssec:results1}.
This enables us to see whether the mapping of these tactile properties for each material in the latent space corresponds to the degree of roughness, hardness, and friction from our calculated values, see~\Cref{fig:visualization}.
The color of the circles enclosing each material number in~\Cref{fig:visualization}~(a) is calculated to be deeper for more rough and harder materials.
For roughness we count the number of oscillations in the z-axis of the tactile sequence, and the absolute maximum/minimum values in the z-axis for hardness (see~\Cref{fig: example of image and tactile}).
These two values are then multiplied to obtain a color value for visualization purposes.
Similarly, the colors of the enclosing circles in~\Cref{fig:visualization}~(b) are based on the amount of friction each material has, again as described in~\Cref{ssec:results1}.
These colors change according to the absolute maximum and minimum values found in the y-axis of the tactile sequences (see~\Cref{fig: example of image and tactile}).
We note that tactile properties are represented in the latent space according to what the tactile sensor perceived.
Therefore, what we perceive as the degree of tactile properties might not correspond to our result.

\Cref{fig:visualization}~(a) indicates that materials with relatively high degree of hardness and roughness tend to get mapped to regions with lower values of the latent variable $z_1$.
For example, material 10 (carpet) was recognized as hard and rough, while material 9 (body towel) and 13 (toilet mat) were recognized as soft and smooth.
Moreover, we see that unknown material 24 (black carpet) is relatively closer to the somewhat similar textured, known material 10 (brown carpet) than to the other unknown materials in the center region, despite their difference in color.
From this point of view, \Cref{fig:visualization}~(a) suggests that the degree of softness and roughness properties of materials are embedded in latent variable $z_1$.
However, we notice that material 5 (bubble wrap) is not mapped properly in this regard.
We believe that although its roughness was obtained from the tactile sensor, its corresponding image features were not obtained due to its transparency.
An interesting case is material 16, which has the covering of a Japanese straw mat surface printed on paper and was estimated to have a high degree of roughness. 
However, tactile values corresponding to this degree of roughness could not be obtained by the sensor.
This shows the limitation of our current model on how accurate tactile properties can be estimated from only two-dimensional images as input.

Furthermore, \Cref{fig:visualization}~(b) indicates that materials with seemingly low friction tend to get mapped to regions with low values of $z_2$.
For example, fabric like materials such as clothing have relatively high friction when stroked by the sensor due to contact with the lycra cover of the tactile sensor.
On the other hand, materials like plastic slip more easily when stroked and have relatively low friction as a result.
We can see that relatively glossy (thus seemingly slippery) materials 1 (table cover), 5 (bubble wrap), and 14 (floor mat) are mapped to areas with the lowest $z_2$ values.
Therefore, we believe that $z_2$ is connected to the amount of friction material surfaces provide during stroking.

%%%%%%%%%%%%%%%%%%%%%%%%%%%%%%%%%%%%%%%%%%%%%%%%%%%%%%%%%%%%%%%%%%%%%%%%%%%%%%%%
\subsection{Comparison with Classification Model}
\begin{figure}
    \centering
	\includegraphics[width=1.0\columnwidth]{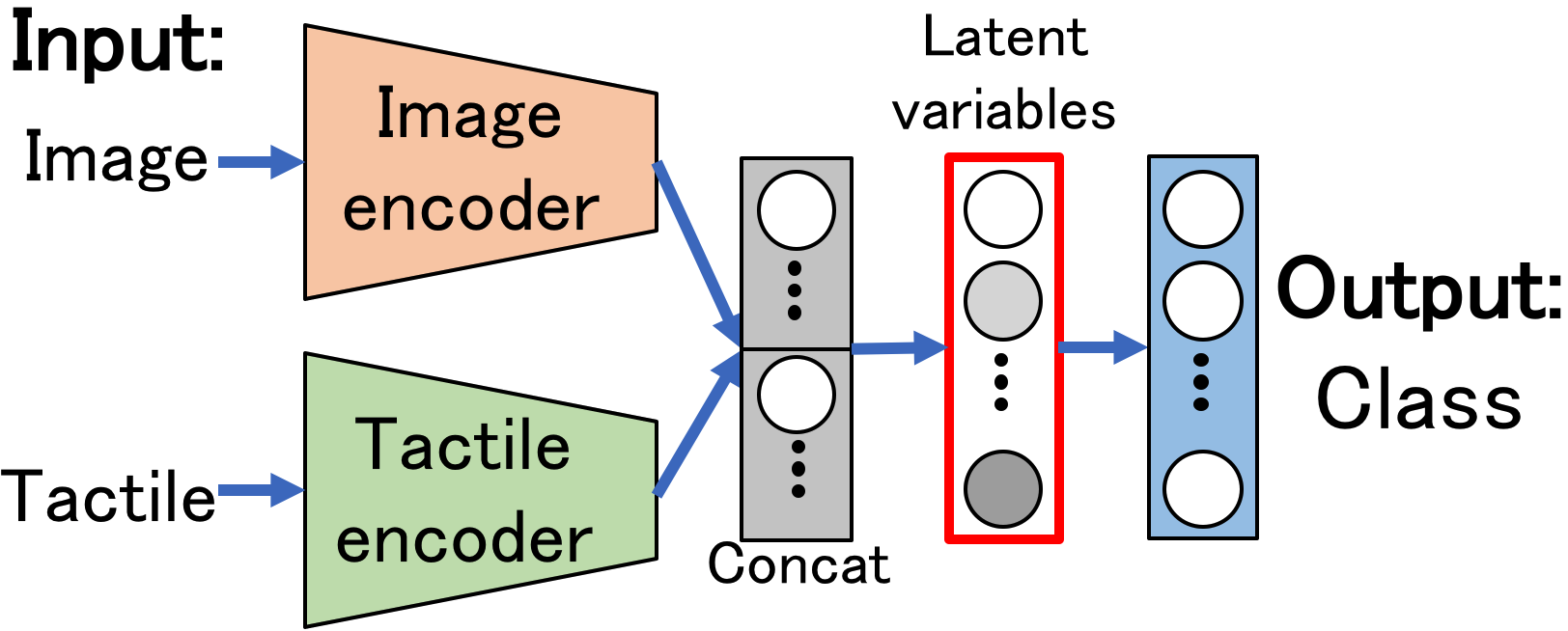}
	\caption{Comparison model composed of two encoder components for image and tactile sequences, hidden layer as latent variables, and classification layer for output.}
	\label{fig:classification_model}
\end{figure}

As comparison against our proposed network $N_{\textsc{proposed}}$, we also create and train a network $N_{\textsc{classification}}$, which outputs classes when given both the image of the material surface and its tactile information as input, see~\Cref{fig:classification_model}.
$N_{\textsc{classification}}$ contains two encoder components; 2D CNN for images, and 3D CNN for tactile sequences. 
Moreover, it has a layer to concatenate the image and tactile features, as well as a hidden layer to mix these features.
Finally, we connect this hidden layer to a softmax layer to perform classification.
Further details on the network parameters are shown in~\Cref{tab:networkdesignclassification}.
Again, we use batch normalization for all layers except last layer in the network.
Furthermore, we used softmax cross entropy as loss function, Adam optimizer with $\alpha=$\num{1e-3} and batch size of 96 for training with the same dataset as in our proposed method.
Training of $N_{\textsc{classification}}$ took about five minutes.

\begin{figure}
    \centering
	\includegraphics[width=1.0\columnwidth]{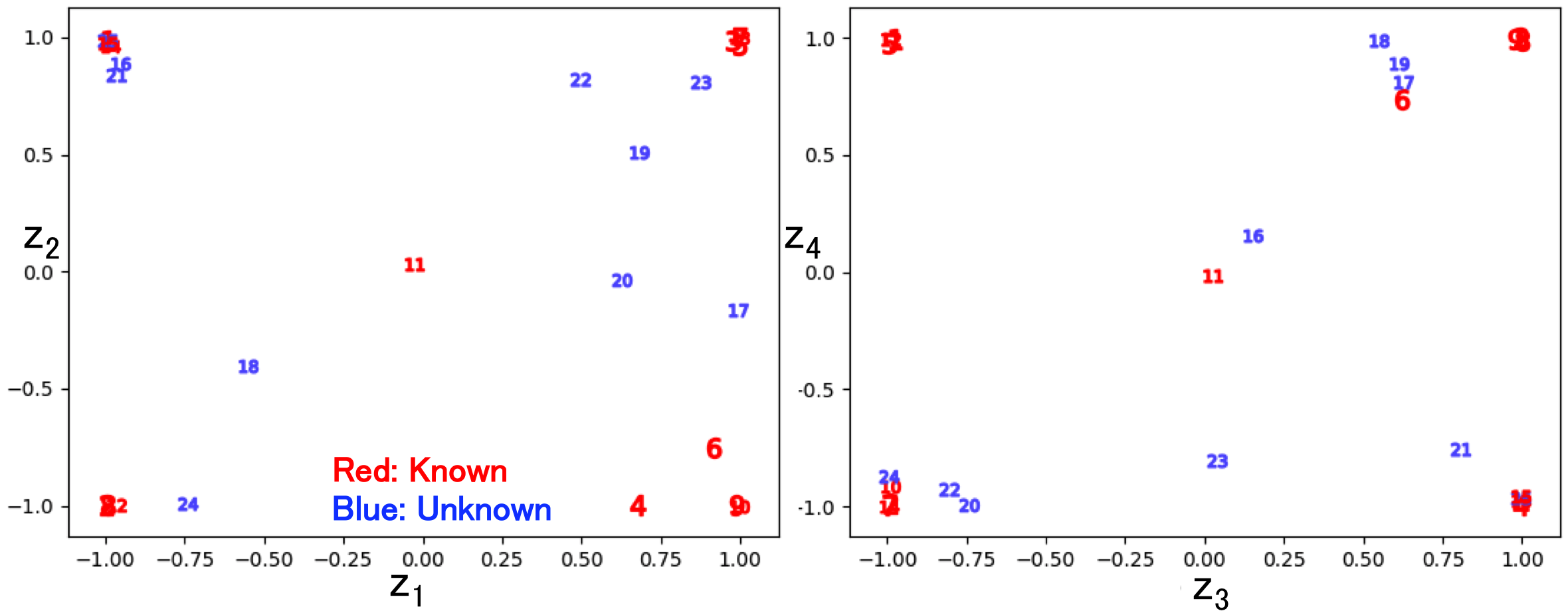}
	\caption{Visualization of latent spaces of the hidden layer from the comparison classification model.}
	\label{fig:classification_result}
\end{figure}

\Cref{fig:classification_result} shows the latent space of $N_{\textsc{classification}}$ with four latent variables.
We see that materials are clearly separated in this latent space when compared to the latent space of $N_{\textsc{proposed}}$, because the output for classes is expressed in a discrete manner.
Despite being represented in continuous space, the latent variables of $N_{\textsc{classification}}$ do not express the degree of tactile properties for each material.
We can conclude that $N_{\textsc{proposed}}$ from our proposed method without classification successfully expresses such levels of tactile properties in continuous space.

\begin{table}
\vspace{5mm}
\centering
\begin{threeparttable}
\caption{Classification Network Design for Comparison\tnote{1}}
\label{tab:networkdesignclassification}
\begin{tabular}{|c|c|c|c|c|c|c|c|}
\hline
 & Layer & In & Out & \shortstack{Filter\\size} & Stride & Padding & \shortstack{Activation\\function}\\
\hline\hline
 \parbox[t]{2mm}{\multirow{4}{*}{\rotatebox[origin=c]{90}{2D Conv.}}}
 		& \nth{1} & 1  & 32 & (8,8) & (2,2) & (0,0) & ReLu\\
        & \nth{2} & 32 & 32 & (8,8) & (2,2) & (0,0) & ReLu \\
        & \nth{3} & 32 & 32 & (4,4) & (2,2) & (0,0) & ReLu \\
        & \nth{4} & 32 & 32 & (4,4) & (2,2) & (0,0) & Tanh \\
\hline
 \parbox[t]{2mm}{\multirow{4}{*}{\rotatebox[origin=c]{90}{3D Conv.}}}
 		& \nth{1} & 3  & 32 & (2,2,4) & (1,1,2) & (0,0,0) & ReLu \\
        & \nth{2} & 32 & 32 & (2,2,4) & (1,1,2) & (0,0,0) & ReLu \\
        & \nth{3} & 32 & 32 & (1,1,3) & (1,1,2) & (0,0,0) & ReLu \\
        & \nth{4} & 32 & 31  & (1,1,3) & (1,1,1) & (0,0,0) & Tanh \\
\hline
\end{tabular}
\begin{tablenotes}
	\item[1] \scriptsize {We use 10 neurons for each image and tactile feature pair in the concat layer, a MLP with two hidden layers connected to the concat layer with 4 neurons and tanh as activation function, and softmax with 15 classes for the classification layer.}
\end{tablenotes}
\end{threeparttable}
\end{table}

%%%%%%%%%%%%%%%%%%%%%%%%%%%%%%%%%%%%%%%%%%%%%%%%%%%%%%%%%%%%%%%%%%%%%%%%%%%%%%%%
%%%%%%%%%%%%%%%%%%%%%%%%%%%%%%%%%%%%%%%%%%%%%%%%%%%%%%%%%%%%%%%%%%%%%%%%%%%%%%%%

\section{Conclusion}
\label{sec:conclusion}
We proposed a method to estimate tactile properties from images, called deep visuo-tactile learning, for which we built an encoder-decoder network with latent variables.
The network is trained with material texture images as input and time series sequences tactile acquired from a tactile sensor as output.
After training, we obtained a continuous latent space representing tactile properties with degrees for various materials.
Our experiments showed that unlike conventional methods relying on classification, our network is able to deal with unknown material surfaces and adapted the latent variables accordingly without the need of manually designed class labels.

For future work, we would like to extend our network to also use 3D images instead of preprocessing the colored images and extracting the edges due to lack of information on reflective surfaces.
%%%%%%%%%%%%%%%%%%%%%%%%%%%%%%%%%%%%%%%%%%%%%%%%%%%%%%%%%%%%%%%%%%%%%%%%%%%%%%%%
%%%%%%%%%%%%%%%%%%%%%%%%%%%%%%%%%%%%%%%%%%%%%%%%%%%%%%%%%%%%%%%%%%%%%%%%%%%%%%%%
%\addtolength{\textheight}{-12cm}
%%%%%%%%%%%%%%%%%%%%%%%%%%%%%%%%%%%%%%%%%%%%%%%%%%%%%%%%%%%%%%%%%%%%%%%%%%%%%%%%

\section*{ACKNOWLEDGMENT} \small
The authors would like to thank Kenta Yonekura for his assistance in making the end-effector for our experiments, and Wilson Ko for proofreading this article.

%%%%%%%%%%%%%%%%%%%%%%%%%%%%%%%%%%%%%%%%%%%%%%%%%%%%%%%%%%%%%%%%%%%%%%%%%%%%%%%%
\bibliographystyle{IEEEtran}
\bibliography{IEEEabrv,bibliography}

\begin{thebibliography}{10}
\providecommand{\url}[1]{#1}
\csname url@rmstyle\endcsname
\providecommand{\newblock}{\relax}
\providecommand{\bibinfo}[2]{#2}
\providecommand\BIBentrySTDinterwordspacing{\spaceskip=0pt\relax}
\providecommand\BIBentryALTinterwordstretchfactor{4}
\providecommand\BIBentryALTinterwordspacing{\spaceskip=\fontdimen2\font plus
\BIBentryALTinterwordstretchfactor\fontdimen3\font minus
  \fontdimen4\font\relax}
\providecommand\BIBforeignlanguage[2]{{%
\expandafter\ifx\csname l@#1\endcsname\relax
\typeout{** WARNING: IEEEtran.bst: No hyphenation pattern has been}%
\typeout{** loaded for the language `#1'. Using the pattern for}%
\typeout{** the default language instead.}%
\else
\language=\csname l@#1\endcsname
\fi
#2}}

\bibitem{bergmann2010tactual}
W.~M. Bergmann-Tiest, ``{Tactual Perception of Material Properties},''
  \emph{Vision Research}, vol.~50, no.~24, pp. 2775--2782, 2010.

\bibitem{tanaka2015investigating}
M.~Tanaka and T.~Horiuchi, ``{Investigating Perceptual Qualities of Static
  Surface Appearance Using Real Materials and Displayed Images},'' \emph{Vision
  Research}, vol. 115, pp. 246--258, 2015.

\bibitem{yanagisawa2015effects}
H.~Yanagisawa and K.~Takatsuji, ``{Effects of Visual Expectation on Perceived
  Tactile Perception: An Evaluation Method of Surface Texture with Expectation
  Effect},'' \emph{International Journal of Design}, vol.~9, no.~1, 2015.

\bibitem{fleming2014visual}
R.~W. Fleming, ``{Visual Perception of Materials and Their Properties},''
  \emph{Vision Research}, vol.~94, pp. 62--75, 2014.

\bibitem{dahiya2013directions}
R.~S. Dahiya, P.~Mittendorfer, M.~Valle, G.~Cheng, and V.~J. Lumelsky,
  ``{Directions Toward Effective Utilization of Tactile Skin: A Review},''
  \emph{IEEE Sensors Journal}, vol.~13, no.~11, pp. 4121--4138, 2013.

\bibitem{ohmura2006conformable}
Y.~Ohmura, Y.~Kuniyoshi, and A.~Nagakubo, ``{Conformable and Scalable Tactile
  Sensor Skin for Curved Surfaces},'' in \emph{IEEE International Conference on
  Robotics and Automation (ICRA)}, 2006, pp. 1348--1353.

\bibitem{iwata2009design}
H.~Iwata and S.~Sugano, ``{Design of Human Symbiotic Robot TWENDY-ONE},'' in
  \emph{IEEE International Conference on Robotics and Automation (ICRA)}, 2009,
  pp. 580--586.

\bibitem{mittendorfer2011humanoid}
P.~Mittendorfer and G.~Cheng, ``{Humanoid Multimodal Tactile-sensing
  Modules},'' \emph{IEEE Transactions on robotics}, vol.~27, no.~3, pp.
  401--410, 2011.

\bibitem{fishel2012sensing}
J.~A. Fishel and G.~E. Loeb, ``{Sensing Tactile Microvibrations with the
  BioTac—Comparison with Human Sensitivity},'' in \emph{IEEE RAS \& EMBS
  International Conference on Biomedical Robotics and Biomechatronics
  (BioRob)}, 2012, pp. 1122--1127.

\bibitem{paulino2017low}
T.~Paulino, P.~Ribeiro, M.~Neto, S.~Cardoso, A.~Schmitz, J.~Santos-Victor,
  A.~Bernardino, and L.~Jamone, ``{Low-cost 3-axis Soft Tactile Sensors for the
  Human-Friendly Robot Vizzy},'' in \emph{IEEE International Conference on
  Robotics and Automation (ICRA)}, 2017, pp. 966--971.

\bibitem{johnson2009retrographic}
M.~K. Johnson and E.~H. Adelson, ``{Retrographic Sensing for the Measurement of
  Surface Texture and Shape},'' in \emph{IEEE Conference on Computer Vision and
  Pattern Recognition (CVPR)}, 2009, pp. 1070--1077.

\bibitem{dong2017improved}
S.~Dong, W.~Yuan, and E.~Adelson, ``{Improved GelSight Tactile Sensor for
  Measuring Geometry and Slip},'' \emph{arXiv preprint arXiv:1708.00922}, 2017.

\bibitem{tomo2016modular}
T.~P. Tomo, W.~K. Wong, A.~Schmitz, H.~Kristanto, A.~Sarazin, L.~Jamone,
  S.~Somlor, and S.~Sugano, ``{A Modular, Distributed, Soft, 3-axis Sensor
  System for Robot Hands},'' in \emph{IEEE-RAS 16th International Conference on
  Humanoid Robots (Humanoids)}, 2016, pp. 454--460.

\bibitem{tomo2018uskin}
T.~P. Tomo, A.~Schmitz, W.~K. Wong, H.~Kristanto, S.~Somlor, J.~Hwang,
  L.~Jamone, and S.~Sugano, ``{Covering a Robot Fingertip With uSkin: A Soft
  Electronic Skin With Distributed 3-Axis Force Sensitive Elements for Robot
  Hands},'' \emph{IEEE Robotics and Automation Letters}, vol.~3, no.~1, pp.
  124--131, Jan 2018.

\bibitem{yuan2017gelsight}
W.~Yuan, S.~Dong, and E.~H. Adelson, ``{GelSight: High-Resolution Robot Tactile
  Sensors for Estimating Geometry and Force},'' \emph{Sensors}, vol.~17,
  no.~12, p. 2762, 2017.

\bibitem{Calandra2017}
R.~Calandra, J.~Lin, A.~Owens, J.~Malik, U.~C. Berkeley, D.~Jayaraman, and
  E.~H. Adelson, ``{More Than a Feeling : Learning to Grasp and Regrasp using
  Vision and Touch},'' no. Nips, pp. 1--10, 2017.

\bibitem{yang2016tactile}
H.~Yang, F.~Sun, W.~Huang, L.~Cao, and B.~Fang, ``{Tactile Sequence Based
  Object Categorization: A Bag of Features Modeled by Linear Dynamic System
  with Symmetric Transition Matrix},'' in \emph{International Joint Conference
  on Neural Networks (IJCNN)}, 2016, pp. 5218--5225.

\bibitem{yamaguchi2016combining}
A.~Yamaguchi and C.~G. Atkeson, ``{Combining Finger Vision and Optical Tactile
  Sensing: Reducing and Handling Errors While Cutting Vegetables},'' in
  \emph{IEEE-RAS International Conference on Humanoid Robots (Humanoids)},
  2016, pp. 1045--1051.

\bibitem{he2016deep}
K.~He, X.~Zhang, S.~Ren, and J.~Sun, ``{Deep Residual Learning for Image
  Recognition},'' in \emph{IEEE conference on computer vision and pattern
  recognition}, 2016, pp. 770--778.

\bibitem{conneau2016very}
A.~Conneau, H.~Schwenk, L.~Barrault, and Y.~Lecun, ``{Very Deep Convolutional
  Networks for Natural Language Processing},'' \emph{arXiv preprint
  arXiv:1606.01781}, 2016.

\bibitem{schmitz2014tactile}
A.~Schmitz, Y.~Bansho, K.~Noda, H.~Iwata, T.~Ogata, and S.~Sugano, ``{Tactile
  Object Recognition Using Deep Learning and Dropout},'' in \emph{IEEE-RAS
  International Conference on Humanoid Robots (Humanoids)}, 2014, pp.
  1044--1050.

\bibitem{baishya2016robust}
S.~S. Baishya and B.~B{\"a}uml, ``{Robust Material Classification With a
  Tactile Skin Using Deep Learning},'' in \emph{IEEE/RSJ International
  Conference on Intelligent Robots and Systems (IROS)}, 2016, pp. 8--15.

\bibitem{yuan2017connecting}
W.~Yuan, S.~Wang, S.~Dong, and E.~Adelson, ``{Connecting Look and Feel:
  Associating the Visual and Tactile Properties of Physical Materials},'' in
  \emph{IEEE Conference on Computer Vision and Pattern Recognition (CVPR17)},
  2017, pp. 21--26.

\bibitem{gao2016deep}
Y.~Gao, L.~A. Hendricks, K.~J. Kuchenbecker, and T.~Darrell, ``Deep learning
  for tactile understanding from visual and haptic data,'' in \emph{2016 IEEE
  International Conference on Robotics and Automation (ICRA)}.\hskip 1em plus
  0.5em minus 0.4em\relax IEEE, 2016, pp. 536--543.

\bibitem{yuan2017shape}
W.~Yuan, C.~Zhu, A.~Owens, M.~A. Srinivasan, and E.~H. Adelson,
  ``{Shape-independent Hardness Estimation Using Deep Learning and a GelSight
  Tactile Sensor},'' in \emph{IEEE International Conference on Robotics and
  Automation (ICRA)}, 2017, pp. 951--958.

\bibitem{bell2015material}
S.~Bell, P.~Upchurch, N.~Snavely, and K.~Bala, ``Material recognition in the
  wild with the materials in context database,'' in \emph{Proceedings of the
  IEEE conference on computer vision and pattern recognition}, 2015, pp.
  3479--3487.

\bibitem{schwartz2017visual}
G.~Schwartz, ``{Visual Material Recognition},'' \emph{Drexel University}, 2017.

\bibitem{ji20133d}
S.~Ji, W.~Xu, M.~Yang, and K.~Yu, ``{3D Convolutional Neural Networks for Human
  Action Recognition},'' \emph{IEEE transactions on pattern analysis and
  machine intelligence}, vol.~35, no.~1, pp. 221--231, 2013.

\bibitem{kingma2014adam}
D.~Kingma and J.~Ba., ``{Adam: A Method For Stochastic Optimization},'' 2015.

\end{thebibliography}
% Aim for 25 references.
\end{document}